\documentclass[conference]{IEEEtran}
\usepackage{cite}
\usepackage{todonotes}
\usepackage{amsmath,amssymb,amsfonts}
\usepackage{algorithm,algpseudocode}

\usepackage{graphicx}
\usepackage{array}
\usepackage{longtable}
\usepackage{booktabs}
\usepackage{comment}      
\usepackage{subcaption}
\usepackage{textcomp}
\usepackage[pagebackref=true,breaklinks=true,letterpaper=true,colorlinks,bookmarks=false]{hyperref}
\usepackage{xcolor}
\def\BibTeX{{\rm B\kern-.05em{\sc i\kern-.025em b}\kern-.08em
    T\kern-.1667em\lower.7ex\hbox{E}\kern-.125emX}}

\begin{document}

\title{Robust Adversarial Defense by Tensor Factorization\\
}

\author{
\IEEEauthorblockN{Manish Bhattarai}
\IEEEauthorblockA{\textit{Theoretical Division, LANL} \\
Los Alamos, USA \\
ceodspspectrum@lanl.gov}
\and
\IEEEauthorblockN{Mehmet Cagri Kaymak \\ Ryan Barron}
\IEEEauthorblockA{\textit{Theoretical Division, LANL} \\
Los Alamos, USA \\}
\and
\IEEEauthorblockN{Ben Nebgen \\ Kim Rasmussen \\ Boian S. Alexandrov}
\IEEEauthorblockA{\textit{Theoretical Division, LANL} \\
Los Alamos, USA \\
}
}

\maketitle 
\begin{abstract}
As machine learning techniques become increasingly prevalent in data analysis, the threat of adversarial attacks has surged, necessitating robust defense mechanisms. Among these defenses, methods exploiting low-rank approximations for input data preprocessing and neural network (NN) parameter factorization have shown potential. Our work advances this field further by integrating the tensorization of input data with low-rank decomposition and tensorization of NN parameters to enhance adversarial defense. The proposed approach demonstrates significant defense capabilities, maintaining robust accuracy even when subjected to the strongest known auto-attacks. Evaluations against leading-edge robust performance benchmarks reveal that our results not only hold their ground against the best defensive methods available but also exceed all current defense strategies that rely on tensor factorizations. This study underscores the potential of integrating tensorization and low-rank decomposition as a robust defense against adversarial attacks in machine learning.

\end{abstract}



\section{Introduction}

The recent advances in machine learning and deep learning have empowered a multitude of applications across various domains, from image recognition to recommender systems. However, these models are susceptible to adversarial attacks, wherein small, carefully crafted perturbations to the input can cause them to output incorrect results\cite{goodfellow2014explaining}. This vulnerability represents a significant challenge to the reliable application of machine learning models, particularly in critical areas such as cybersecurity~\cite{rosenberg2021adversarial} and healthcare~\cite{finlayson2019adversarial}.  This raises two primary concerns: the impact on the credibility of current machine learning systems and the danger of malevolent adversarial attacks in real-world applications.

Motivated  by the above-mentioned challenges, our study seeks to enhance our understanding of adversarial attacks and provide efficient and robust defense mechanisms against them. We are specifically interested in exploring the capabilities of tensor factorization as a means to defend against adversarial incursions in the image domain. We propose effective defenses that can be incorporated without significantly altering the original model structure or performance. We leverage an extensive parameter search for tensor factorization method to counter attacks, with a focus on the preservation of core data features in the process of eliminating adversarial perturbations.

\begin{table}[ht!]
\centering
\begin{tabular}{|c|c|c|c|}
\hline
\textbf{Dataset (Metric, $\epsilon$)} & \textbf{Method} & \textbf{Clean} & \textbf{AA} \\ \hline
CIFAR-10 $(l_{\infty}, \epsilon = 8/255)$ & Rank \#1 &  \textbf{93.25} & \textbf{70.69} \\ \cline{2-4} 
                   
                          & \textbf{Ours}      & 85.59 & 70.24 \\ \hline
CIFAR-10 $(l_2, \epsilon = 128/255)$ & Rank \#1  &   \textbf{95.54} & \textbf{84.86} \\ \cline{2-4}  
                       
                            & \textbf{Ours}      & 86.61 & 77.73 \\ \hline
CIFAR-100 $(l_{\infty}, \epsilon = 8/255)$ & Rank \#1 & \textbf{75.22} & 42.67 \\ \cline{2-4} 
                        
                           & \textbf{Ours}      & 60.12 & \textbf{42.68} \\ \hline
\end{tabular}
\caption{Comparison of test accuracy(\%) from our tensorial denoiser to the state-of-the-art model, as in RobustBench~\cite{croce2021robustbench}.}
\label{compare}
\end{table}

\section{Related Work}
\label{sec:relevant_work}
Tensor decompositions as a defense against adversarial attacks on deep learning (DL) models were first introduced in \cite{cho2020applying} to demonstrate a simple yet effective approach to resist the attacks without significant degradation compared to the model's original performance on clean data.  Another work, ``defensive tensorization'', presented in \cite{bulat2021defensive}, proposed a novel adversarial defense technique that employed a latent high-order factorization of network layers, then applied tensor dropout in the latent subspace to yield dense reconstructed weights. This work demonstrated effective versatility across multitudinous domains, reaching low-precision architectures. Entezari and Papalexakis investigated adversarial attacks on recommendation systems to present defense methods such as low-rank reconstructions as well as a transformation of attacked data. \cite{entezari2022low}

Block-term Dropout (BT-Dropout) was proposed in \cite{ouyang2022block}  whereby a network was factorized into a latent high-order representation, imposing a low-rank block-term tensor structure on the weights of the fully-connected layer. They applied BT-Dropout in the latent subspace without directly pruning the weights.  He et al. \cite{he2021tdnn} presented an adversarial defense method based on tensor decomposition (TDNN), which decomposed then reconstructed images to maintain critical features. This process removed adversarial example perturbations, exhibiting improved defense along with lower run times relative to traditional tensor decomposition.

 Tensor layers plus tensor dropout implemented in convolutional neural networks (CNNs) as a means to improve inductive bias, robustness, and efficiency by using low-rank tensor structures on the weights of tensor regression layers was introduced in Ref. \cite{kolbeinsson2021tensor}. This work established superior performance post-model-modification, furthering robustness against noise and adversarial attacks. 

Samangouei et al. \cite{samangouei2018defense} presented Defense-GAN. This new framework utilized generative models' capabilities to protect deep neural networks against adversarial attacks by modeling the distribution of unperturbed images paired with the removal of adversarial changes during inference. Similarly, Ilyas et. al  \cite{ilyas2019adversarial} posits that adversarial examples in machine learning are due to non-robust features expressed as predictive data distribution patterns but fragile for model performance, even unintuitive to humans. These features, identified within a theoretical framework and shown to be prevalent in standard datasets, reflect a disconnect between human-defined concepts of robustness due to inherent data geometry.

Current tensor-based denoising frameworks lack strategies for selecting optimal hyper-parameters, like tensor ranks for optimal decomposition, and their defense mechanisms are tested only against relatively simple attacks such as FGSM and PGD. Our work addresses this gap, inspired by previous research on low-rank approximation tensor decomposition defense strategies. Our approach combines the low-rank approximated tensorized  image, which reshapes into a higher-order tensor by  stacking image patches, and the reparametrization of the neural networks (NN) with tensors. We leverage both Tucker~\cite{kolda2009tensor} and Tensor-Train-based~\cite{oseledets2011tensor} decompositions and highlight the efficacy of our optimal rank selection strategy. Furthermore, to facilitate direct comparisons with existing state-of-the-art defense frameworks, we assess our model's effectiveness using AutoAttack~\cite{croce2020reliable}, a well-recognized standard in the field.  This innovative approach enhances the robustness and efficiency of the denoising process, thereby improving the overall system performance.

\section{Background}
This section discusses our defense model's mathematical foundation.  Our denoiser model, as illustrated in Figure~\ref{img1}, is comprised of patch extract, patch merge, and tensor factorization.

\subsection{Tensorization of images}

\subsubsection{Patch Extraction}

Given an input image as $I \in \mathbb{R}^{C \times W \times H}$, where $C$ is the number of channels, $W$ is the width, and $H$ is the height. Similarly, the kernel size (patch size) is represented as $K$, the stride as $S$, the padding as $P$, and the dilation as $D$. With the above defined, the Image $I$ is tensorized into a 4-D tensor $O \in \mathbb{R}^{\frac{W-K+2P}{S}\times \frac{H-K+2P}{S} \times C \times K \times K}$, where each $K \times K$ patch in the input tensor becomes a column in the output tensor as:

\begin{figure}[ht]
    \centering
    \includegraphics[width=.9\linewidth]{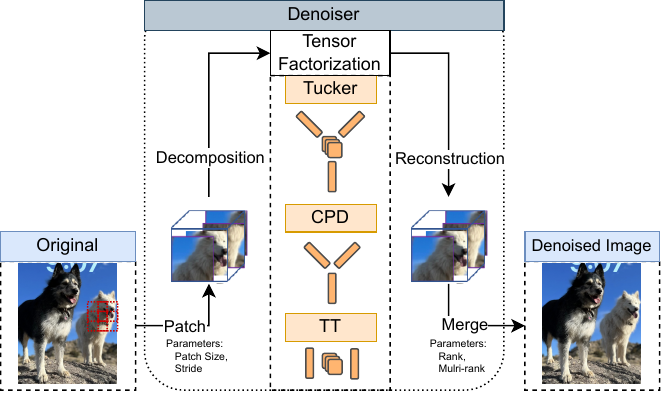}
    \caption{Overview of the Tensorial denoiser}
    \label{img1}
\end{figure}

\begin{align}
O_{ w, h,c, k_1, k_2} & = I_{c, S \cdot w + D \cdot k_1, S \cdot h + D \cdot k_2} \label{eqn1} \\ 
& \forall k_1, k_2 \in [0, K] \nonumber \\
& \forall w \in [0, \frac{W-K+2P}{S}] \nonumber \\
& \forall h \in [0, \frac{H-K+2P}{S}] \nonumber  
\end{align}

Here, $O_{ w, h,c, k_1, k_2}$ is the pixel value at the position $(k_1, k_2)$ in the patch at position $(w, h)$ for the channel $c$ in the output tensor. $I_{c, S \cdot w + D \cdot k_1, S \cdot h + D \cdot k_2}$ is the corresponding pixel value in the input tensor. $s$ is the stride and $d$ is the dilation.

\begin{figure*}[ht]
    \centering
    \includegraphics[width=.8\textwidth]{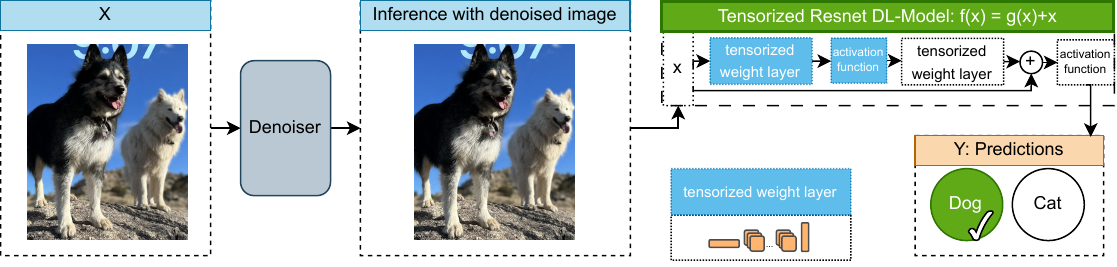}
    \caption{Overview of the Adversarial denoising setup}
    \label{img2}
\end{figure*}

\subsubsection{Patch Merge}
This operation does the inverse of the operation performed in patch extraction, folding the patches back to their original locations to reform the image. Mathematically, this operation can be expressed as follows:

\begin{align}
I_{c, w, h} & = \frac{1}{C_{w, h}} \sum_{k_1=0}^{K} \sum_{k_2=0}^{K} O_{ \frac{w - D \cdot k_1}{S}, \frac{h - D \cdot k_2}{S},c, k_1, k_2}\\
& \quad \cdot \mathbb{I}_{ S \mid (w - D \cdot k_1), S \mid (h - D \cdot k_2) } \nonumber \\
& \quad \forall w \in [0, W], h \in [0, H], \nonumber
 \label{eqn:2}
\end{align}

where

\begin{align}
C_{w, h} & = \sum_{k_1=0}^{K} \sum_{k_2=0}^{K} \mathbb{I}_{ S \mid (w - D \cdot k_1), S \mid (h - D \cdot k_2)} \\
& \quad \forall w \in [0, W], h \in [0, H].\nonumber
 \label{eqn:3}
\end{align}

Here, $I_{c,w,h}$ is the resulting image, $O_{ w, h, c, k_1, k_2}$ are the patches to be folded back into the image, and $\mathbb{I}{ \cdot }$ is the indicator function that ensures the conditions inside the brackets hold. The function returns 1 if the condition is true and 0 otherwise.

The indicator function $\mathbb{I}_{ S \mid (w - D \cdot k_1), S \mid (h - D \cdot k_2) }$ ensures the original patch extraction positions in the image are correctly restored. The positions $w$ and $h$ are chosen to be multiples of the stride $S$ and are positioned correctly for the dilation $D$ and the patch indices $k_1$ and $k_2$. This patch and merge strategies are shown in Figure~\ref{img1}.

\subsection{Tensor Decomposition}
Once the image is transformed into patches, we perform a low-rank approximation of the tensors with tensor decomposition. We apply Tucker Decomposition~\cite{kolda2009tensor} and Tensor train decomposition~\cite{oseledets2011tensor} in this work. The details about these methods are presented in the following sections.

\subsubsection{Tucker Decomposition}
Given the tensor patch $O \in \mathbb{R}^{M \times N \times C \times K \times K }$, the Tucker decomposition is:

\begin{equation}
O \approx \mathcal{G} \times_1 A^{(1)} \times_2 A^{(2)} \times_3 A^{(3)} \times_4 A^{(4)} \times_5 A^{(5)},
\end{equation}

where $\mathcal{G} \in \mathbb{R}^{R_1 \times R_2 \times R_3 \times R_4 \times R_5}$ is the core tensor, $A^{(n)} \in \mathbb{R}^{I_n \times R_n}$ are the factor matrices for each mode ($n = 1,2,3, 4, 5$), and $\times_n$ denotes the $n$-mode product.

\subsubsection{Tensor Train Decomposition}
Tensor Train decomposition, given the tensor patch $O \in \mathbb{R}^{ M \times N \times C \times K \times K}$, is as follows:

\begin{align}
O(i_1, i_2, i_3, i_4, i_5) & \approx \sum_{r_1, r_2, r_3, r_4} G^{(1)}_{i_1, r_1} G^{(2)}_{r_1, i_2, r_2} \\
& G^{(3)}_{r_2, i_3, r_3} G^{(4)}_{r_3, i_4, r_4} G^{(5)}_{r_4, i_5} \nonumber,
\end{align}

where $G^{(n)}$ are the TT cores for each mode, and the indices $r_n$ (called ranks) represent the connections between the cores~\cite{oseledets2011tensor}.

The low-rank, compressed representation of the patched image tensor is then reconstructed using the Patch Merge algorithm. The decompressed image, where high-frequency noise has been removed through tensor factorization, is then classified with the DL model. An overview of utilizing this denoiser model for adversarial denoising is shown in Figure~\ref{img2}.

\subsection{Tensorizing Neural Network}

In addition to applying tensor factorizations as preprocessing steps for input images, we also execute a low-rank re-parameterization on the NN layers. While low-rank approximation of NNs was initially designed for significantly reducing the parameters yielding great accelerations~\cite{lee2019learning,kim2015compression,astrid2017cp,lebedev2014speeding,novikov2015tensorizing}, they have also demonstrated promising results for adversarial defense\cite{he2021tdnn,bulat2021defensive}. In an NN with a convolutional kernel parameterized as $\mathbf{S}\in\mathbb{R}^{d\times d\times P\times Q}$, where $d$ is the filter size, and $P$ and $Q$ are the number of input and output channels respectively, the weight tensor can be factorized using the Tucker Decomposition. The factorization is expressed as:

\begin{equation}
    \tilde{\mathbb{S}}_{i,j,p,q} = \sum_{r_p=1}^{R_p} \sum_{r_q=1}^{R_t} \mathbb{G}_{i,j,r_p,r_q} \mathbb{A}^P_{p,r_p}\mathbb{A}^Q_{q,r_q}
\end{equation}

Here, $\mathbb{G}$ is the reduced kernel tensor, $R_p$ and $R_q$ are the ranks of input and output feature map dimensions, respectively, and $\mathbb{A}^P$ and $\mathbb{A}^Q$ are the factor matrices corresponding to the input and output feature maps. This factorization transforms a single convolutional layer into three distinct convolutional layers: two ($1\times 1$) layers for $\mathbb{A}^P$ and $\mathbb{A}^Q$, and a $d\times d$ convolutional layer for $\mathbb{G}$. Consequently, the complexity of the original layer is significantly reduced, leading to potential computational and storage efficiency. 

Alongside Tucker, we employ tensor-train based reparameterization of the NN. Here, the TT factorization is given as: 

\begin{equation}
    \tilde{\mathbb{S}}_{i,j,p,q} = \sum_{r_1=1}^{R_1}\sum_{r_2=1}^{R_2}\sum_{r_3=1}^{R_3} \mathbb{G}^1_{i,r_1} \mathbb{G}^2_{r_1,j,r_2} \mathbb{G}^3_{r_2,p,r_3} \mathbb{G}^4_{r_3,q}
\end{equation}

Each $\mathbb{G}^k$ is a TT-core, and $R_k$ are the TT-ranks. This factorization transforms the 4D tensor $\mathbb{S}$ into a sequence of matrices and vectors, which can be stored and manipulated more efficiently. The TT-decomposition is mainly used on high-order ($>3$) tensors, providing efficient tensor operations yet more compression than the Tucker decomposition. 
\begin{figure}[ht!]
    \centering
    \includegraphics[width=1\linewidth]{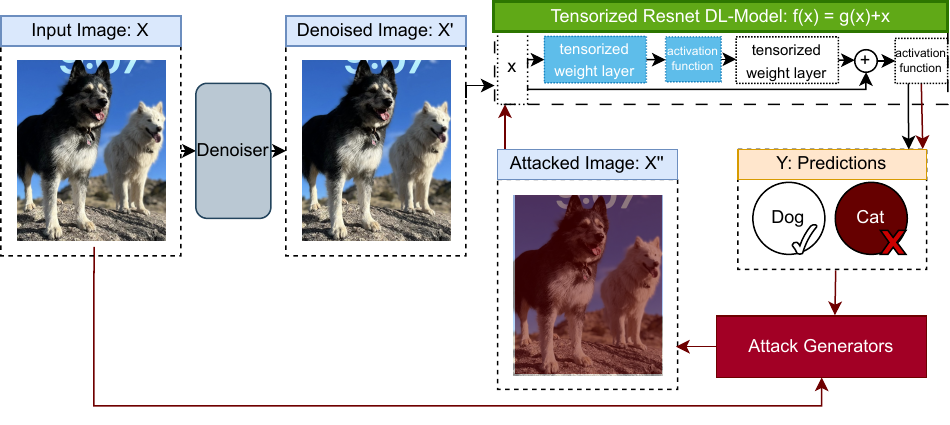}
    \caption{Overview of the attack generation}
    \label{img3}
\end{figure}

\subsection{Adversarial Attacks}

The recent advancements in adversarial defense frameworks are commonly evaluated for their robustness using an ensemble of adversarial attacks bundled in a tool called AutoAttack~\cite{croce2020reliable}. We, too, utilize this tool to assess the effectiveness of our proposed defense mechanism. AutoAttack integrates white-box and black-box attack methodologies tailored to challenge a model's performance under adversarial conditions. The white-box attack methods include Auto-PGD (Projected Gradient Descent) and APGD-DLR (Auto PGD with Difference of Logits Ratio). In contrast, the black-box attack techniques encompass FAB (Fast Adaptive Boundary attack) and Square Attack.

Auto-PGD and APGD-DLR constitute gradient-based approaches. These methods strategically leverage gradient data to iteratively adjust inputs until a  misclassification results. They utilize constraints by limiting the alterations within a predefined threshold, denoted as $\epsilon$, governed by the attack norms ($l_1$ or $l_{\infty}$). On the other hand, FAB and Square Attack are decision-based attacks that generate adversarial examples by making small, $\epsilon$-bounded adjustments to the input. These modifications are fine-tuned to observe whether they lead to misclassification by the model.
AutoAttack enables the generation of adversarial samples that probe the performance of our denoising model under severe black-box and white-box adversarial attacks. This helps us determine how our framework fares when subjected to these rigorous tests. The overview of the attack generation process is shown in Figure~\ref{img3}. Initially, the adversarial attack is crafted for one DL model, such as ResNet or WideResNet. Once generated, these adversarial samples are processed through the denoiser before passage to the DL model.

\begin{figure*}[ht!]
    \centering
    \includegraphics[width=.8\textwidth]{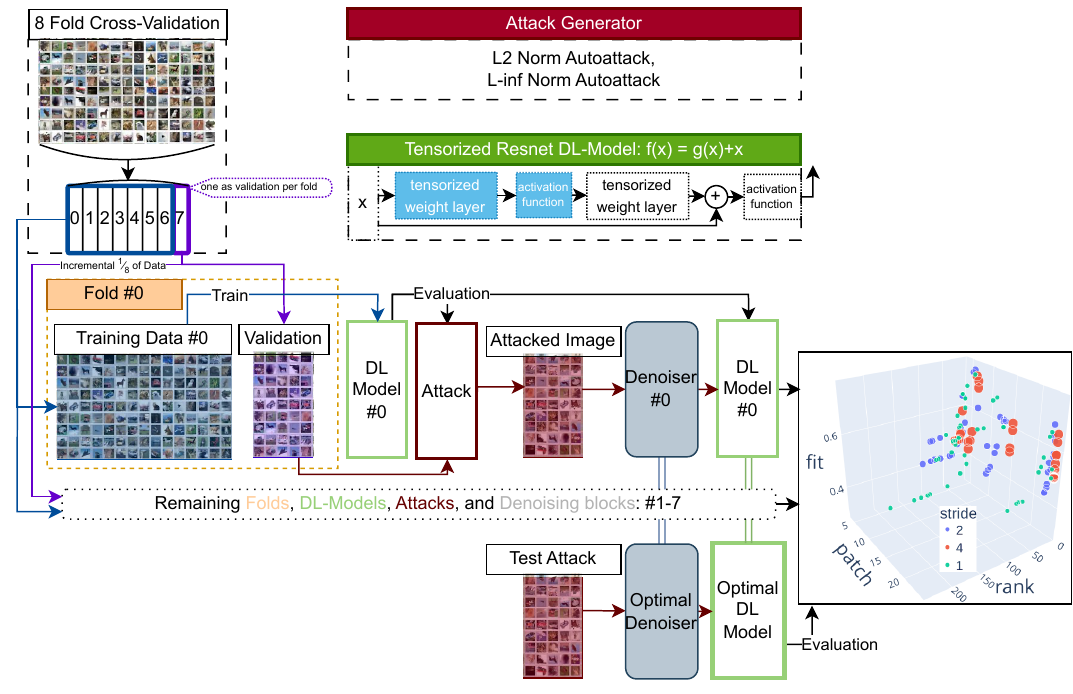}
         \caption{Overview of comprehensive training and inference pipeline encapsulating the process of optimal parameter selection through eight-fold cross-validation, adversarial attack generation using Autoattack, the employment of the denoiser block equipped with optimized hyperparameters, and the final tensor decomposition for computational efficiency and robustness. }

    \label{fig:overview_final}
   
\end{figure*}

\section{Method}
\label{sec:method}

Figure~\ref{fig:overview_final} offers a comprehensive overview of our training and inference pipeline. We use an eight fold cross-validation approach to optimally select the parameters for the denoising block, such as \textit{patch size}, \textit{stride}, and \textit{tensor decomposition ranks}. The training data is divided into eight partitions. Each training fold consists of seven partitions used for training, leaving one partition for validation. This process results in eight distinct training partitions, thus yielding eight trained DL models. While we utilized CIFAR10 and CIFAR100 datasets for the above experiments, we employed Resnet18 and WideResnet28-10 DL models for adversarial robustness evaluation. We utilized the PyTorch Lightning framework with appropriate seeding  to ensure reproducibility across experiments. We set the learning rate of $1e-2$, batch size of $256$, and maximum training epoch of $200$. The experiments were run on NVIDIA A100 GPUs of 80GB memory. For hyperparameter tuning, we exploited Ray Tune\cite{liaw2018tune} framework for handling parallel workloads. 

Autoattack is subsequently employed to create attacks corresponding to the validation sets for each model. For a one-to-one comparison with state-of-the-art adversarial robustness framework, we exploited $L_1$ and $L_{\infty}$ attacks with perturbation $\epsilon$ equals to $8/255$ and $128/255$ respectively. Almost all the datasets subjected to Autoattack resulted in 0 evaluation accuracy for the DL model.  

The generated attacks are then fed into the denoiser block. Here, the block is evaluated using different combinations of hyperparameters such as \textit{patch size}, \textit{stride}, and \textit{tensor decomposition ranks}. These hyperparameters were selected using an optimization technique known as "Tree-Structured Parzen Estimator"~\cite{bergstra2011algorithms} specifically implemented through Optuna~\cite{akiba2019optuna}, a Python library for hyperparameter optimization. Specifically, we have four hyperparameters to tune: patch\textunderscore size, stride, $\textrm{rank}_{p}$, and $\textrm{rank}_{k}$. patch\textunderscore size is the size of the patches into which the image is divided, and stride determines the overlap between these patches. These parameters are selected from categorical lists of potential values, specifically [4, 8, 16, 24] for patch size and [1, 2, 4] for stride. Although a patch is 2D, the same  rank ($\textrm{rank}_{p}$) is chosen to control the tensor decomposition's rank along both patch axes to simplify the search space for tucker decomposition. Lastly, parameter $\textrm{rank}_{k}$ for tucker decomposition controls the decomposition's rank along the number of patches. The rank corresponding to the channel dimension is kept as it is. Our approach is defined-by-run, meaning the hyperparameter search space is dynamically constructed during the optimization process since the feasible ranges for $\textrm{rank}_{k}$ and $\textrm{rank}_{p}$ are determined by the patch size and stride parameters (i.e., the decomposition rank cannot be larger than the size of the corresponding tensor dimension). A step size of 4 is used for the rank hyperparameters to prune the search space further. The ranks' maximum and minimum allowed values are carefully calculated based on the patch size and stride, ensuring an optimal balance between computational efficiency and performance. For the TT decomposition, the hyperparameters $\textrm{rank}_{k}$ and $\textrm{rank}_{p}$ have a different interpretation. They determine the sizes of the "connecting" dimensions in the train of tensors (TT-cores), controlling the complexity of the multilinear relationships between the tensor dimensions. Building upon Equation~\ref{eqn1}, let's consider a tensorized image, $O$, in its 5D form. The first two dimensions of this tensor are contracted to form a simpler 4D tensor, denoted as $O \in \mathbb{R}^{\frac{(W-K+2P)(H-K+2P)}{S^2} \times C \times K \times K}$. For the Tucker decomposition process, the multi-ranks are arranged in the configuration $[\textrm{rank}_{k},3,\textrm{rank}_{p},\textrm{rank}_{p}]$. In contrast, the  TT ranks  were configured as $[1,\textrm{rank}_{k}, \textrm{rank}_{p}, 3, 1]$ for the reshaped tensor $O \in \mathbb{R}^{\frac{(W-K+2P)(H-K+2P)}{S^2} \times K \times K \times C}$. This reshaping of the tensor is required for the appropriate selection of TT ranks.

The specific configuration of the DL model and the corresponding attacks on the validation dataset determine the optimal parameters for the denoiser. The denoiser is evaluated based on a fitness score metric, the average clean and adversarial accuracy. Then, the optimal denoiser is selected based on the parameter configuration that yields the highest fitness score across the independent folds. For the DL model, we adopt the model that is trained on the entire training set. Early stopping based on validation loss is employed to find the optimal model. Once we determine the optimal denoiser and DL model, an attack is launched on the test set and evaluated using the denoiser.

Moreover, we also incorporate low-rank approximation of the neural network weights to accelerate the DL model and provide additional robustness across adversarial attacks. We exploited Tucker/TT based tensor decomposition to compress the weight matrices of NN without significant loss of useful information. The reduced complexity of the model aided in quicker processing and efficient memory usage, accelerating the model's performance. Simultaneously, low-rank approximation also helped in minimizing the impact of adversarial perturbations as it aids in capturing the dominant, most important features of the data while potentially discarding minute, less meaningful perturbations that are commonly used in adversarial attacks. The tensor decomposition was performed on the final DL model, where the ranks for decomposition were achieved using the Bayesian approach presented in \cite{nakajima2013global}.

In summary, our comprehensive framework offers a robust and efficient solution for DL models facing adversarial attacks.  The solution is achieved through cross-validation to select optimal parameters, a denoiser block for robustness against adversarial attacks, and low-rank approximations for computational efficiency. This approach sets a solid foundation for future studies to enhance the resilience and performance of DL models in adversarial scenarios.

\section{Results}
\label{sec:results}
\begin{figure}[ht!]
    \centering
    \includegraphics[width=1\linewidth]{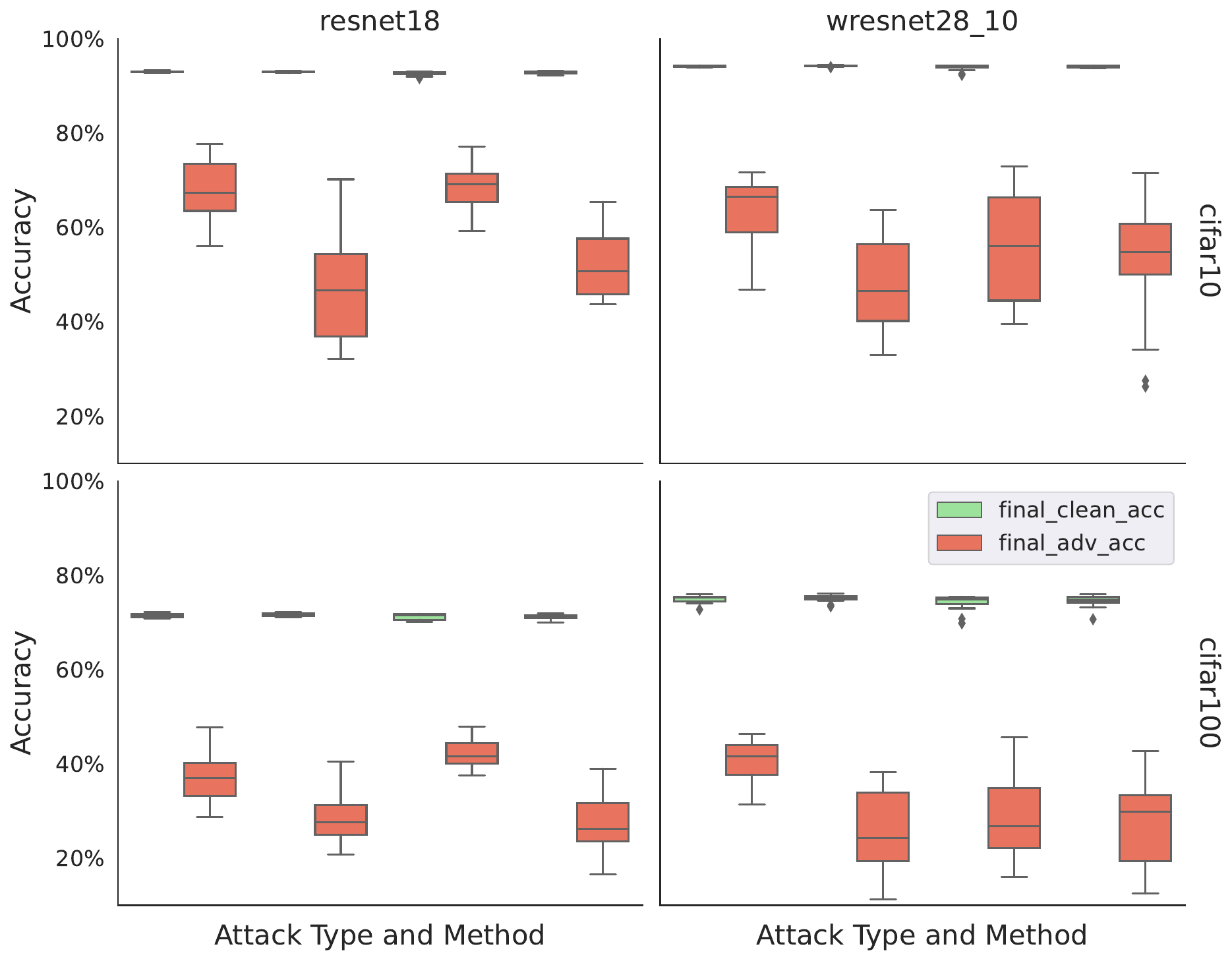}
    \caption{Distribution of clean and adversarial accuracy scores achieved for  top 10 denoiser hyperparameter configurations for test dataset.}
    \label{img6}
\end{figure}

Figure~\ref{img6} presents the comprehensive performance of our cross-validated model when evaluated against the test dataset. The first 10 denoiser hyperparameter setups, which deliver the maximum test performance, are specifically reported. Both clean and adversarial accuracy scores are provided, following the procedure of passing the images through the denoiser and feeding them into the Deep Learning (DL) models. 

\begin{figure}[ht!]
    \centering
    \includegraphics[width=1\linewidth]{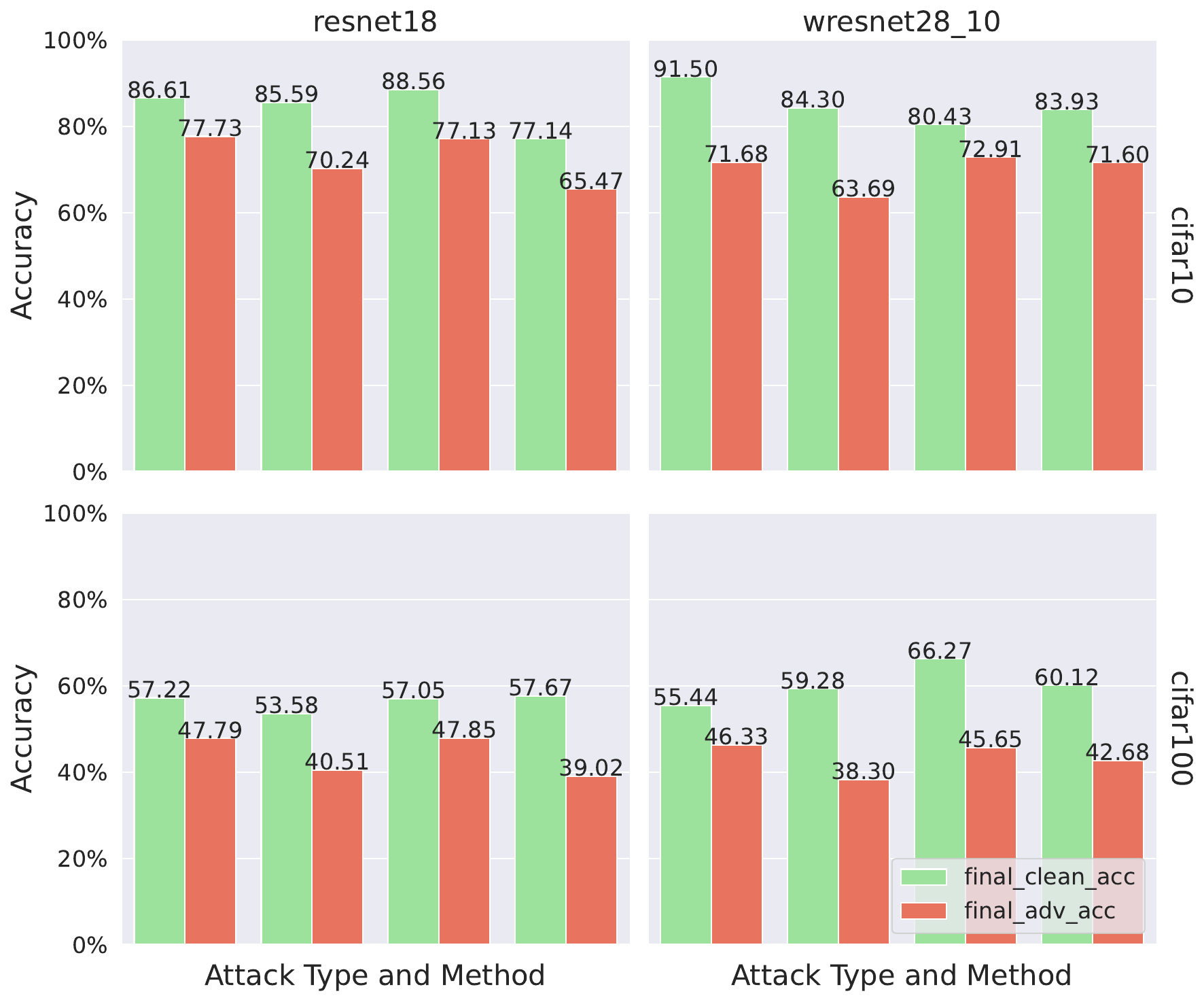}
    \caption{Statistical representation of adversarial metrics corresponding to the optimal hyperparameter configuration that maximizes the average of clean and adversarial accuracy.}
    \label{img5}
\end{figure}
The clean accuracy results demonstrate remarkable consistency, as indicated by the minimal variance in the accuracy plot. In contrast, the adversarial accuracy shows considerable variability. However, relative high-performance levels are maintained, indicating our denoising model's effective and consistent ability to eliminate adversarial perturbations. Detailed performance scores can be found in tables \ref{tab:cifar10_wres},\ref{tab:cifar100_wres},\ref{tab:cifar10_res18}, and \ref{tab:cifar100_res18}, where we provide a summary of the hyperparameters leading to the optimal fitness score. Tables \ref{tab:cifar10_wres} and \ref{tab:cifar100_wres} provide the results of our denoising model when applied to the CIFAR10 and CIFAR100 datasets, respectively, using the WideResnet18-10 model. Meanwhile, tables \ref{tab:cifar10_res18} and \ref{tab:cifar100_res18} present the denoising model's performance on the same datasets but using the Resnet18 model instead. In all scenarios, the models and datasets were evaluated using the AutoAttack method with an $L_{\infty}$ norm of $8/255$ and an $L_2$ norm of $128/255$. We present the denoising model's performance under these configurations (DL model, datasets, attack norm), employing Tensor Train and Tucker tensor decomposition methods.

The results obtained from various model configurations highlight the influence of different adversarial attack scenarios. For instance, when the WideResnet architecture is employed on the CIFAR10 dataset, with an $L_{\infty}$ attack and Tucker decomposition (as detailed in table \ref{tab:cifar10_Linf_tucker}), the configuration yielding the highest adversarial accuracy (0.7160) involved a patch size of 24, a stride of 1, a $\textrm{rank}_{k}$ of 60, and a $\textrm{rank}_{p}$ of 12. Conversely, when the Resnet18 architecture was used with the CIFAR100 dataset, subjected to an $L_2$ attack and TT decomposition (refer table \ref{tab:cifar100_L2_tt}), the top-performing model configuration (achieving an adversarial accuracy of 0.4633) had a patch size of 8, a stride of 2, a $\textrm{rank}_{k}$ of 32, and a $\textrm{rank}_{p}$ of 12. These variances in performance across different model configurations underscore the diverse impacts of adversarial attacks. Furthermore, in Figure~\ref{img5}, we present the robustness metric, highlighting the hyperparameters corresponding to the highest clean and adversarial accuracy average.

\begin{figure*}[ht!]
    \centering
        \begin{subfigure}{.3\textwidth}
        \centering
        \includegraphics[width=\linewidth]{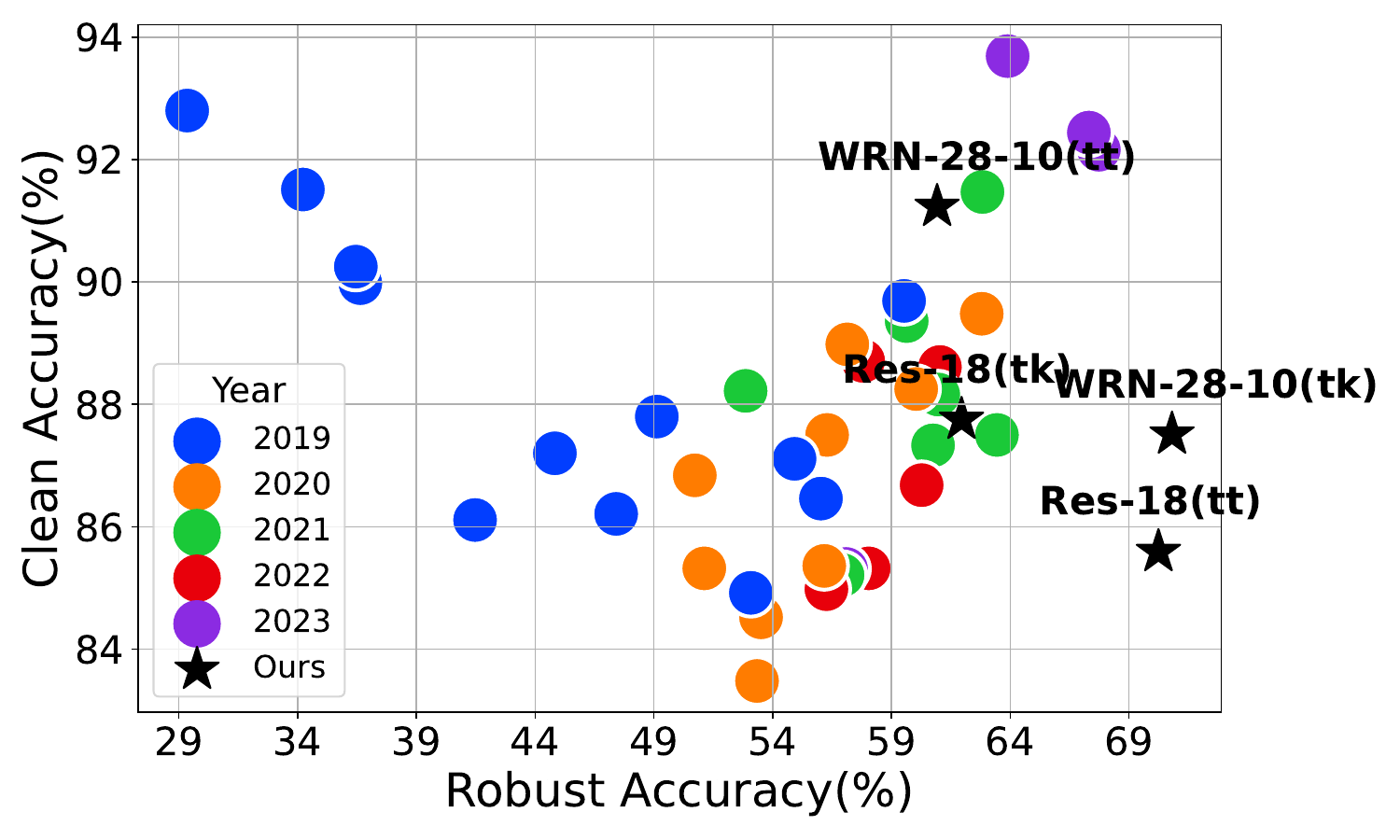}
        \caption{CIFAR-10($l_{\infty}$,$\epsilon = \frac{8}{255})$}
        \label{fig:ima1}
    \end{subfigure}
        \begin{subfigure}{.3\textwidth}
        \centering
        \includegraphics[width=\linewidth]{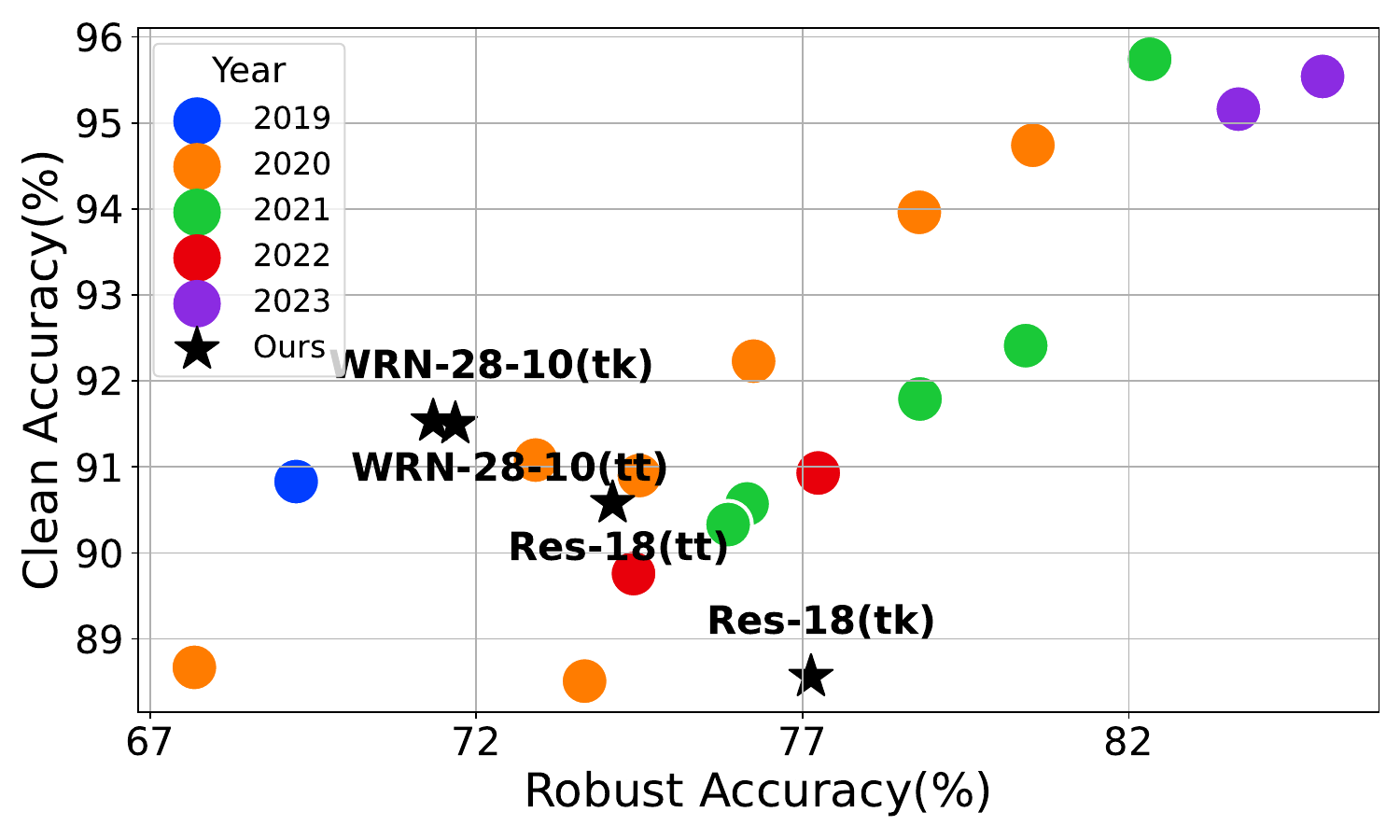}
        \caption{CIFAR-10($l_{2}$,$\epsilon = \frac{128}{255}$)}
        \label{fig:ima2}
    \end{subfigure}
        \begin{subfigure}{.3\textwidth}
        \centering
        \includegraphics[width=\linewidth]{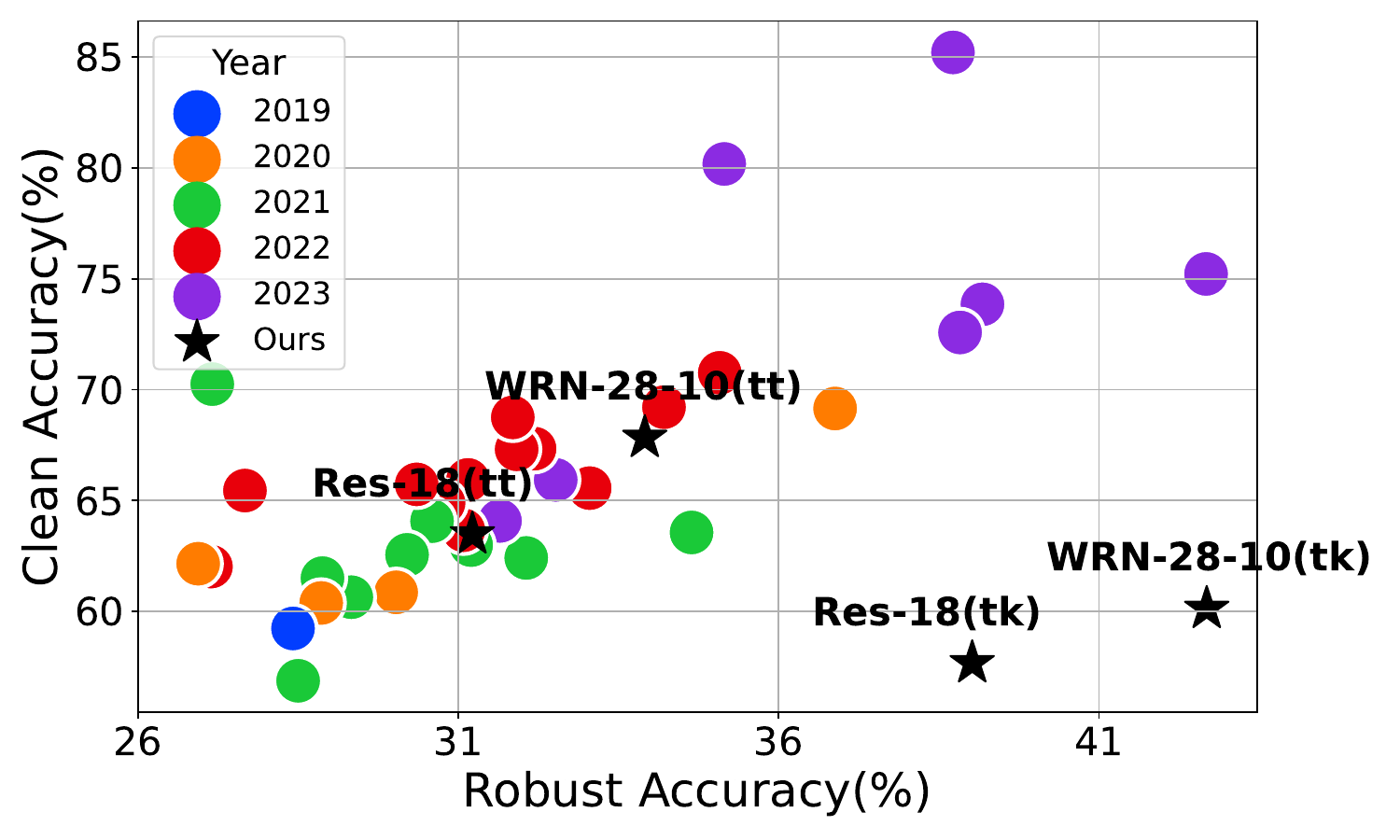}
        \caption{CIFAR-100($l_{\infty}$,$\epsilon = \frac{8}{255}$)}
        \label{fig:ima3}
     \end{subfigure}
     \caption{Robust accuracy against AutoAttack and clean accuracy of top-rank models from RobustBench~\cite{croce2021robustbench}. Different colors are used to indicate top-rank models for different publication years. Our denoising model was based on Tucker(tk) and Tensor-Train(tt) for Resnet-18(Res-18) and WideResnet-70-16 (WRN-70-16) architectures. }
     \label{fig:final_compare}
\end{figure*}

We have also compared our findings with the current best-in-class adversarial robust models, using the Robustbench benchmark~\cite{croce2021robustbench}. This comparison is depicted in Figure~\ref{fig:final_compare}, and Table~\ref{compare} presents a detailed contrast of our top-performing denoising model against the state-of-the-art adversarial robustness model that is based on denoising diffusion~\cite{wang2023better}.

For the CIFAR-10 dataset, the leading adversarial robust framework achieves a top accuracy of 93.25\% on clean data and 70.69\% on an AutoAttack dataset with $L_{\infty}$ norm. In contrast, our top model performs comparably with 85.59\% and 70.24\% accuracy, respectively.  Unlike most existing denoising models presented, our approach does not require additional datasets for training or any form of adversarial training.  This may be one of the reasons why the clean accuracy of our approach trails behind the existing defense frameworks. Notably, our model surpasses the top-ranked model on the CIFAR-100 dataset, outperforming it by a margin of  0.01\% on the attacked dataset for the $L_{\infty}$ norm. However, our model does not perform as well under an $L_2$ AutoAttack, trailing behind the best model by a margin of approximately 7\%. This shortfall could be attributed to utilizing the Tensor Train (TT) and Tucker tensor decomposition tools with an $L_2$ minimization objective, which may inadvertently allow for adversarial noise during reconstruction. To address this issue, leveraging the Tensor decomposition tool based on $L_1$ minimization ~\cite{chachlakis2019l1,liu2003robust} may improve resistance against adversarial noise.

Despite its simplicity, our denoising model demonstrates competitive performance compared to the best existing models. The most significant advantage of our model over others lies in its real-time denoising capabilities.



\begin{table}[t]
\centering
\caption{Top 5 Final Adversarial Accuracies for cifar10 and Wide-Resnet}
\label{tab:cifar10_wres}
\begin{subtable}{.5\textwidth}
\centering
\caption{Linf Attack, Tucker Decomposition}
\label{tab:cifar10_Linf_tucker}
\begin{tabular}{cccccc}
\toprule
patch & stride & $\textrm{rank}_{k}$ & $\textrm{rank}_{p}$  & clean acc & adv acc \\
\midrule
24 & 1 & 60 & 12 & 0.8393 & 0.7160 \\
8 & 2 & 24 & 20 & 0.8751 & 0.7081 \\
8 & 4 & 20 & 8 & 0.8029 & 0.6941 \\
8 & 1 & 36 & 8 & 0.9075 & 0.6164 \\
24 & 1 & 36 & 20 & 0.7159 & 0.6148 \\
\bottomrule
\end{tabular}
\end{subtable}
\hfill
\begin{subtable}{.5\textwidth}
\centering
\caption{Linf Attack, TT Decomposition}
\label{tab:cifar10_Linf_tt}
\begin{tabular}{cccccc}
\toprule
patch & stride & $\textrm{rank}_{k}$ & $\textrm{rank}_{p}$  & clean acc & adv acc \\
\midrule
4 & 1 & 12 & 8 & 0.8430 & 0.6369 \\
8 & 1 & 44 & 8 & 0.9123 & 0.6092 \\
8 & 1 & 40 & 16 & 0.9125 & 0.5815 \\
24 & 1 & 44 & 16 & 0.8533 & 0.5722 \\
16 & 1 & 64 & 20 & 0.7609 & 0.5681 \\
\bottomrule
\end{tabular}
\end{subtable}
\newline
\vspace{10pt}
\begin{subtable}{.5\textwidth}
\centering
\caption{L2 Attack, Tucker Decomposition}
\label{tab:cifar10_L2_tucker}
\begin{tabular}{cccccc}
\toprule
patch & stride & $\textrm{rank}_{k}$ & $\textrm{rank}_{p}$  & clean acc & adv acc \\
\midrule
8 & 1 & 52 & 8 & 0.8043 & 0.7291 \\
8 & 2 & 28 & 8 & 0.9150 & 0.7134 \\
4 & 2 & 12 & 12 & 0.7921 & 0.7118 \\
8 & 1 & 32 & 12 & 0.8640 & 0.7078 \\
8 & 1 & 40 & 20 & 0.8951 & 0.6791 \\
\bottomrule
\end{tabular}
\end{subtable}
\hfill
\begin{subtable}{.5\textwidth}
\centering
\caption{L2 Attack, TT Decomposition}
\label{tab:cifar10_L2_tt}
\begin{tabular}{cccccc}
\toprule
patch & stride & $\textrm{rank}_{k}$ & $\textrm{rank}_{p}$  & clean acc & adv acc \\
\midrule
8 & 1 & 48 & 8 & 0.8215 & 0.7182 \\
8 & 2 & 32 & 16 & 0.8893 & 0.7026 \\
4 & 1 & 12 & 12 & 0.7820 & 0.6923 \\
24 & 1 & 40 & 12 & 0.8367 & 0.6801 \\
8 & 1 & 36 & 16 & 0.9137 & 0.6674 \\
\bottomrule
\end{tabular}
\end{subtable}
\end{table}

\begin{table}[t]
\centering
\caption{Top 5 Final Adversarial Accuracies for cifar100 and Wide-Resnet}
\label{tab:cifar100_wres}
\begin{subtable}{.5\textwidth}
\centering
\caption{Linf Attack, Tucker Decomposition}
\label{tab:cifar100_Linf_tucker}
\begin{tabular}{cccccc}
\toprule
patch & stride & $\textrm{rank}_{k}$ & $\textrm{rank}_{p}$  & clean acc & adv acc \\
\midrule
8 & 1 & 40 & 8 & 0.6012 & 0.4268 \\
8 & 1 & 32 & 12 & 0.4787 & 0.3444 \\
8 & 2 & 36 & 8 & 0.4111 & 0.3430 \\
24 & 1 & 44 & 20 & 0.6322 & 0.3429 \\
4 & 2 & 12 & 8 & 0.6650 & 0.3422 \\
\bottomrule
\end{tabular}
\end{subtable}
\hfill
\begin{subtable}{.5\textwidth}
\centering
\caption{Linf Attack, TT Decomposition}
\label{tab:cifar100_Linf_tt}
\begin{tabular}{cccccc}
\toprule
patch & stride & $\textrm{rank}_{k}$ & $\textrm{rank}_{p}$  & clean acc & adv acc \\
\midrule
8 & 4 & 28 & 8 & 0.5928 & 0.3830 \\
8 & 1 & 40 & 12 & 0.6131 & 0.3510 \\
8 & 1 & 36 & 12 & 0.6661 & 0.3445 \\
8 & 2 & 48 & 8 & 0.6650 & 0.3432 \\
4 & 1 & 12 & 12 & 0.6690 & 0.3396 \\
\bottomrule
\end{tabular}
\end{subtable}
\newline
\vspace{10pt}
\begin{subtable}{.5\textwidth}
\centering
\caption{L2 Attack, Tucker Decomposition}
\label{tab:cifar100_L2_tucker}
\begin{tabular}{cccccc}
\toprule
patch & stride & $\textrm{rank}_{k}$ & $\textrm{rank}_{p}$  & clean acc & adv acc \\
\midrule
8 & 2 & 32 & 12 & 0.6627 & 0.4565 \\
8 & 1 & 40 & 20 & 0.5770 & 0.4217 \\
8 & 1 & 52 & 8 & 0.6761 & 0.4176 \\
8 & 1 & 32 & 16 & 0.4953 & 0.3644 \\
8 & 1 & 40 & 12 & 0.7079 & 0.3551 \\
\bottomrule
\end{tabular}
\end{subtable}
\hfill
\begin{subtable}{.5\textwidth}
\centering
\caption{L2 Attack, TT Decomposition}
\label{tab:cifar100_L2_tt}
\begin{tabular}{cccccc}
\toprule
patch & stride & $\textrm{rank}_{k}$ & $\textrm{rank}_{p}$  & clean acc & adv acc \\
\midrule
8 & 2 & 32 & 12 & 0.5544 & 0.4633 \\
8 & 1 & 40 & 12 & 0.5521 & 0.4600 \\
8 & 4 & 32 & 8 & 0.6630 & 0.4563 \\
4 & 1 & 12 & 8 & 0.5594 & 0.4478 \\
8 & 1 & 44 & 20 & 0.6690 & 0.4470 \\
\bottomrule
\end{tabular}
\end{subtable}
\end{table}

\begin{table}[t]
\centering
\caption{Top 5 Final Adversarial Accuracies for cifar10 and Resnet18}
\label{tab:cifar10_res18}
\begin{subtable}{.5\textwidth}
\centering
\caption{Linf Attack, Tucker Decomposition}
\label{tab:cifar10_Linf_tucker}
\begin{tabular}{cccccc}
\toprule
patch & stride & $\textrm{rank}_{k}$ & $\textrm{rank}_{p}$  & clean acc & adv acc \\
\midrule
8 & 2 & 16 & 8 & 0.7714 & 0.6547 \\
8 & 1 & 36 & 8 & 0.8775 & 0.6195 \\
8 & 1 & 20 & 8 & 0.7724 & 0.6060 \\ 
8 & 2 & 28 & 20 & 0.6583 & 0.5935 \\
8 & 1 & 28 & 8 & 0.8122 & 0.5782 \\

\bottomrule
\end{tabular}
\end{subtable}
\hfill
\begin{subtable}{.5\textwidth}
\centering
\caption{Linf Attack, TT Decomposition}
\label{tab:cifar10_Linf_tt}
\begin{tabular}{cccccc}
\toprule
patch & stride & $\textrm{rank}_{k}$ & $\textrm{rank}_{p}$  & clean acc & adv acc \\
\midrule
8 & 4 & 16 & 8 & 0.8559 & 0.7024 \\
8 & 1 & 24 & 8 & 0.7960 & 0.6788 \\
24 & 2 & 24 & 12 & 0.8572 & 0.6147 \\
8 & 2 & 20 & 24 & 0.7779 & 0.5818 \\
8 & 4 & 12 & 16 & 0.8100 & 0.5739 \\
\bottomrule
\end{tabular}
\end{subtable}
\newline
\vspace{10pt}
\begin{subtable}{.5\textwidth}
\centering
\caption{L2 Attack, Tucker Decomposition}
\label{tab:cifar10_L2_tucker}
\begin{tabular}{cccccc}
\toprule
patch & stride & $\textrm{rank}_{k}$ & $\textrm{rank}_{p}$  & clean acc & adv acc \\
\midrule
8 & 1 & 28 & 16 & 0.8856 & 0.7713 \\
16 & 1 & 72 & 20 & 0.8899 & 0.7469 \\
16 & 1 & 68 & 20 & 0.8756 & 0.7212 \\
16 & 1 & 80 & 16 & 0.8479 & 0.7184 \\
8 & 2 & 32 & 24 & 0.8798 & 0.7177 \\
\bottomrule
\end{tabular}
\end{subtable}
\hfill
\begin{subtable}{.5\textwidth}
\centering
\caption{L2 Attack, TT Decomposition}
\label{tab:cifar10_L2_tt}
\begin{tabular}{cccccc}
\toprule
patch & stride & $\textrm{rank}_{k}$ & $\textrm{rank}_{p}$  & clean acc & adv acc \\
\midrule

8 & 1 & 36 & 20 & 0.8661 & 0.7773 \\
8 & 2 & 32 & 24 & 0.8408 & 0.7633 \\
8 & 1 & 44 & 8 & 0.9058 & 0.7409 \\
8 & 2 & 32 & 12 & 0.8552 & 0.7398 \\
8 & 4 & 16 & 12 & 0.8087 & 0.7362 \\
\bottomrule
\end{tabular}
\end{subtable}
\end{table}

\begin{table}[t]
\centering
\caption{Top 5 Final Adversarial Accuracies for cifar100 and Resnet18}
\label{tab:cifar100_res18}
\begin{subtable}{.5\textwidth}
\centering
\caption{Linf Attack, Tucker Decomposition}
\label{tab:cifar100_Linf_tucker}
\begin{tabular}{cccccc}
\toprule
patch & stride & $\textrm{rank}_{k}$ & $\textrm{rank}_{p}$  & clean acc & adv acc \\
\midrule
24 & 1 & 32 & 16 & 0.5767 & 0.3902 \\
4 & 1 & 12 & 8 & 0.4781 & 0.3498 \\
24 & 1 & 56 & 16 & 0.4526 & 0.3399 \\
8 & 4 & 16 & 8 & 0.6160 & 0.3250 \\
8 & 4 & 16 & 12 & 0.5181 & 0.3232 \\
\bottomrule
\end{tabular}
\end{subtable}
\hfill
\begin{subtable}{.5\textwidth}
\centering
\caption{Linf Attack, TT Decomposition}
\label{tab:cifar100_Linf_tt}
\begin{tabular}{cccccc}
\toprule
patch & stride & $\textrm{rank}_{k}$ & $\textrm{rank}_{p}$  & clean acc & adv acc \\
\midrule
8 & 2 & 32 & 12 & 0.5358 & 0.4051 \\
116 & 1 & 68 & 24 & 0.5785 & 0.3617 \\
8 & 1 & 40 & 24 & 0.5106 & 0.3554 \\
24 & 2 & 20 & 12 & 0.5636 & 0.3169 \\
24 & 2 & 24 & 12 & 0.4393 & 0.3158 \\
\bottomrule
\end{tabular}
\end{subtable}
\newline
\vspace{10pt}
\begin{subtable}{.5\textwidth}
\centering
\caption{L2 Attack, Tucker Decomposition}
\label{tab:cifar100_L2_tucker}
\begin{tabular}{cccccc}
\toprule
patch & stride & $\textrm{rank}_{k}$ & $\textrm{rank}_{p}$  & clean acc & adv acc \\
\midrule
8 & 1 & 32 & 8 & 0.5705 & 0.4785 \\
24 & 1 & 48 & 16 & 0.6255 & 0.4769 \\
8 & 1 & 36 & 16 & 0.6132 & 0.4629 \\
8 & 2 & 24 & 24 & 0.6028 & 0.4502 \\
4 & 1 & 12 & 8 & 0.6015 & 0.4444 \\
\bottomrule
\end{tabular}
\end{subtable}
\hfill
\begin{subtable}{.5\textwidth}
\centering
\caption{L2 Attack, TT Decomposition}
\label{tab:cifar100_L2_tt}
\begin{tabular}{cccccc}
\toprule
patch & stride & $\textrm{rank}_{k}$ & $\textrm{rank}_{p}$  & clean acc & adv acc \\
\midrule

8 & 1 & 40 & 24 & 0.5722 & 0.4779 \\
4 & 2 & 12 & 12 & 0.4992 & 0.4373 \\
16 & 1 & 64 & 20 & 0.6088 & 0.4310 \\
8 & 1 & 32 & 24 & 0.5572 & 0.4274 \\
8 & 4 & 24 & 8 & 0.6600 & 0.4089 \\
\bottomrule
\end{tabular}
\end{subtable}
\end{table}

\section{Conclusion}
\label{sec:conclusion}
This study has presented a comprehensive evaluation of a Tensor factorization based denoising model and its impact on the performance of deep learning models under various adversarial attack scenarios. Tensor Train and Tucker tensor decompositions have demonstrated noteworthy results with different configurations of deep learning models, datasets, and adversarial attack norms.
Our analysis, encompassing several robustness metrics, has revealed the efficacy of our denoising model under both clean and adversarial conditions. Notably, we have achieved consistent results under clean conditions while effectively mitigating adversarial perturbations, as evidenced by the considerable variability in adversarial accuracy. 
The benchmarking of our denoising model against state-of-the-art adversarial robust models revealed that our model's performance is competitive, even exceeding the performance of top-ranked models in some scenarios. In particular, our model outperformed the leading model on the CIFAR-100 dataset under $L_{\infty}$ norm AutoAttack. 
Our findings indicate the potential of the proposed denoising model to significantly enhance the robustness of deep learning models against adversarial attacks.


\section{Acknowledgment}
This manuscript has been assigned LA-UR-23-27984. This research was funded by the Los Alamos National Laboratory (LANL) Laboratory Directed Research and Development (LDRD) program under grant 20230287ER and supported by LANL's Institutional Computing Program, and by the U.S. Department of Energy National Nuclear Security Administration under Contract No. 89233218CNA000001.

\bibliographystyle{IEEEtran}
\bibliography{main}

\end{document}